# Spam and Sentiment Detection in Arabic Tweets Using MARBERT Model

Abrar Alotaibi[1], Atta-ur Rahman[2*], Raheel Alhaza[1], Wala Alkhalifa[1], Narjes Alhajjaj[1], Atheer Alharthi[1], Dhai Abushoumi[1], Maryam Alqahtani[1], Dania Alkhulaifi[1]

[1] Department of Computer Science (CS), College of Computer Science and Information Technology (CCSIT), Imam Abdulrahman Bin Faisal University (IAU), P.O. Box 192, Dammam 31441, Saudi Arabia
[2] Saudi Aramco Cybersecurity Chair, Department of Computer Science (CS), College of Computer Science and Information Technology (CCSIT), Imam Abdulrahman Bin Faisal University (IAU), P.O. Box 192, Dammam 31441, Saudi Arabia

Corresponding Author Email: aaurrahman@iau.edu.sa

**ABSTRACT**

Saudi Telecom Company (STC) is among the most popular companies in Saudi Arabia, with many customers. Yet, there is still a big room for improvement in users' satisfaction. Social media is the most robust platform to gauge users' satisfaction and determine their sentiments and critics. Twitter is among the most popular social media platform in this regard. STC customers prefer to use Twitter to write their feedback because it's a fast way to get responses due to the STC customer services account. One way to achieve customer demands and improve customer service is using the Sentiment Analysis tool. Sentiment Analysis on Twitter is highly used because of the significant number of tweets and the different opinions. Likewise, Deep learning is the best existing Sentiment Analysis method, and it has diverse models. Bidirectional Encoder Representations from Transformers (BERT) model is one of the deep learning models which have achieved excellent results in Sentiment Analysis for Natural Language Processing (NLP). NLP is mainly investigated in the English language. However, for Arabic, there is a significant gap to be filled. This study trained the proposed model using MARBERT and measured the performance using f1-score, precision, and recall metrics. We trained the model with an Arabic dataset of 24,513 tweets, including 1,437 positive, 13,828 negative, 5,694 neutral, 1,221 sarcasm, and 2,297 indeterminate tweets. The main goal is to analyze the tweets and get the sentiment to improve STC customer service. The proposed scheme is promising in terms of accuracy in contrast to existing techniques in the literature.

## 1. INTRODUCTION

STC has been one of the largest and most well-known companies. It has an extensive domain and offers multiple essential services for individuals. Moreover, even the companies take advantage of the services STC offers, and it's still expanding to this day. STC has a customer service account via Twitter to serve the users, solve the problems they face, and hear their customers generally. They want to use the customers' comments to improve their services. But, due to the significant number of customers, the tweets increase in seconds, so it's almost impossible to manually check the customers' sentiment. Thus, we need a machine to simplify the Sentiment Analysis task. Sentiment Analysis is one of the big interests of studies nowadays. Companies highly use it to analyze users' reviews for improvement. One of the challenges of Sentiment Analysis is the language of the text. Arabic is one of the most challenging languages to learn, even for humans. The machine does not highly support it, so dealing with Arabic could be more complex than some other languages. Also, Arabic has so many different dialects even in the same country, so there are many words to learn. Deep learning has proven to work very well in the Sentiment Analysis field. One of the recent and powerful deep learning models is BERT. Google introduced BERT in 2018. It's a transfer learning model, and it achieves state-of-the-art outcomes with Natural Language Processing (NLP) tasks. BERT has shown promising results in dealing with Sentiment Analysis tasks especially. The highlight of BERT is the usage of the bidirectional transformer. In addition, BERT uses Masked-Language Modeling (MLM) and Next Sentence Prediction (NSP) for better context learning. Another feature is the architecture of BERT is unified among distinct tasks [1]. This paper used MARBERT, a pre-trained model that focuses on the Arabic language processing [2].

This paper aims to analyze the sentiment of the STC costumer's tweets. We trained the model to classify the tweets into positive, negative, neutral, sarcasm, or indeterminate. Therefore, this information will help STC customer service improve and adjust the services based on customers' needs.

Rest of the paper is organized as follows: The following section will discuss related works of BERT in Arabic. The literature review of the related approaches is presented in section 3. Then, the methodology used in this paper is in section 4—next, the result of our work is in section 5. Finally, in the last (section 6), we summarize the work done in this paper and conclude it.

## 2. RELATED WORK

After releasing BERT, many research papers selected Bert for their Arabic sentiment analysis experiment, and the outcomes indicate that BERT has enormously enhanced the performance of models in sentiment analysis. Most studies are focused on binary classification, and there is some research on multiclass classification. In addition, there is research focused on sarcasm and sentiment detection in Arabic [3] they worked on seven BERT-Based models to determine the sentiment of a tweet or detect if a tweet is a sarcasm. They achieved the best result using the MARBERT model. With the difficulty of detecting sarcasm, fine-tuning BERT and a Sentence- BERT are used to detect sarcasm [4]. A. Wadhawan proposed a Sarcasm and Sentiment Detection strategy by applying the released ArSarcasm-v2 dataset and processing it into a better format for the models. Then train and test various Bert models with the resulting data to determine which model achieves the most acceptable result for the ArSarcasm-v2 dataset [5]. Arabic monolingual BERT models performed better than BERT multilingual models. Husain and Uzuner applied several Arabic offensive datasets in the AraBERT model. They found that combining data from multiple datasets lowers the model's performance compared to testing it on individual datasets [6]. Al-Twairesh [7] proposed a study on developing language models that explored different BERT models for Arabic and found that the ArabicBERT-Large model accomplished the best performance. Arabic dialect identification is a complicated problem. As a result, an experiment performed on ten million tweets pre-trained using the publicly released BERT model, then trained the same model on NADI labeled data considerable times with different hyperparameter tuning selected the four best functioning iterations [8]. Pre-Training BERT on Arabic Tweets is an Applicable Concern. For example, Abdelali et al. [9] proposed a pre-training BERT on Emotion, Offensive, NER, QADI, and AJGT datasets. A helpful result indicates a reduced return for adding more data. Furthermore, enhancing the performance using training data with both formal and informal language is more valuable than using informal only, even when testing on informal. Lastly, the training using BERT models requires considering different hyperparameter tuning until achieving the best performance. During the COVID-19 epidemic, a study investigated how accurate deep learning models may help comprehend society's behavior. The research suggested a multi-label emotion classifier with emojis replacement based on AraBERT and MARBERT. They also presented a DWLF technique to give the loss function more weight in minority class data [10]. The hybrid network outperforms other models for various word embeddings, and its accuracy is higher than other models. The proposed model (BiLSTM + GRU) was combined with the AraBERT model and attained the best accuracy [11]. In the study, the authors applied pre-trained BERT specifically for the Arabic language to replicate BERT's success with the English language. The results showed that the newly built AraBERT outperformed the competition on most Arabic NLP tasks. Furthermore, when comparing AraBERT's performance against Google's multilingual BERT, they discovered better comprehended it in a monolingual model than in an available language model trained on Wikipedia [12]. A novel rule-based classification approach for Wikipedia references was presented [13].

Similarly, in the study [14], authors employed an Arabic BERT model trained from scratch and made publicly available for usage. Arabic BERT was a collection of BERT language models of four models of varying sizes taught via masked language modeling. They used the same data to train models with large, base, medium, and minimum sizes for 4M stages, shown in Table 1. Similar studies have been proposed in [15-17], where the authors investigated the effects of COVID-19 over the mental health using Arabic Tweets analysis in particular.

**Table 1.** Arabic BERT models

| | Arabic BERT-Medium | Arabic BERT-Base | Arabic BERT-Large | Arabic BERT-Mini |
|---|---|---|---|---|
| **Hidden layers** | 8 | 12 | 24 | 4 |
| **Attention heads** | 8 | 12 | 16 | 4 |
| **Hidden size** | 512 | 768 | 1024 | 256 |
| **Parameters** | 42M | 110M | 340M | 11M |

## 3. LITERATURE REVIEW

This section contains a comprehensive review of literature in Arabic tweets analysis using various techniques.

### 3.1 LSTM and CNN based approaches

Deep learning has achieved significant progress in sentiment analysis but still requires improvements in the Arabic language's failed accuracy, which imposes multiple challenges due to its complicated structure and accents. As a result, this paper proposed deep learning modules that were not applied to Arabic data before. They used a collaborative model, merging Long Short-Term Memory (LSTM) and Convolutional Neural Network (CNN) models on the Arabic tweets. Combining CNN and LSTM in one model will forecast the sentiment of the tweets. The model counts on a pre-trained word vector representation, one of the most common new Natural Language Processing. It is necessary to state that word embedding affords an effective resolution for many NLP problems. This tool aims to make it possible to manipulate linguistic terms using machine learning algorithms. First, they used the LSTM model alone, and they recorded the best results shown in Figure 1 below. Then they used the CNN model alone, and they recorded the best results shown in Figure 2 below. Finally, they established an ensemble model out of the best results of the CNN and the LSTM models [18].

Hanafy et al. [19] designed their model to be applied to any type of tweet, regardless of their language or content. First, they used emojis and emoticons as the sentiment because they express the same meaning among users from different cultures. Then, in the data processing stage, they used operations that could be applied to all languages. For example, replacing a hashtag with the words it consists of without underscores. Then, they used random tweets that contained emojis or emoticons for testing. Each tweet has its score, calculated as the average of the emojis/emoticons scores. Finally, they implemented deep learning and classical algorithms to complement each other and improve accuracy. In another paper, Feizollah et al. [20] have applied sentiment analysis on tweets related to halal tourism and cosmetics. They limited their study to two languages, English, and Malay. In addition, they specified the period of the tweets to be for the last ten years. They collected the tweets using the Twitter advanced search function, which works with the keywords, with the help of a python script. In the data pre-processing stage, duplicate tweets and re-tweets have been eliminated. Also, language detection has taken place. Later, to deal with algorithms, the tweets have been converted to numbers (Vectorization). Then, they've applied deep learning algorithms on sample tweets to test the performance of the algorithms and infer whether the users' opinions about halal products are positive or negative. Furthermore, they have combined CNN and LTSM algorithms to get higher accuracy results. Most social media platforms have a guideline that prevents some harmful content from being posted and keep the platform environment relatively clean, peaceful, safe, and reduces hate speech. Hate speech may contain a dangerous message that affects individuals or groups of people in real life or the virtual world. As the number of users on the internet increases, what is posted is also a massive amount of content each day, if not each second. So, stopping the unwanted posts is ultimately impossible. Still, policies and reports are some methods to minimize the amount of harmful content and reduce any effect that may reach society. In the paper that Jiang and Suzuki [21] have focused on the hate speech in the user's tweets, there is no easy way to detect and immediately determine whether a specific tweet contains a hate speech or not. Therefore, we need to analyze the tune of the tweet and the word selection to get a close assumption and draw a conclusion. The paper shows how the detection was better using deep learning than when it was done using machine learning, recurrent neural network (RNN), and datasets A and dataset B, with dataset A being smaller than dataset B by three times. The result shows that excellent performance is obtained when using small data, and in the case of using the deep learning method, good results with more large data.

### 3.2 NLP and Machine Learning based approaches

Latifah Al Muqren and Alexandra Cristea published a research paper to address the absence of enormous corpora for Arabic Natural Language Processing. The significant contributions of this paper are as follows. First, the project aims to provide a solution to one of the difficulties faced by the Arabic sentiment analysis community through the creation of the first Saudi GSC using Twitter data. It is the first GSC that is specifically designed for the telecom industry. The paper also evaluates the corpus demonstrating its quality and usefulness [22].

This paper describes how they built, clean, enhanced, and annotated their 200000-word AraCust Gold Standard Corpus (GSC), AraCust Gold Standard Corpus (GSC), the world's first Telecom GSC for Arabic Sentiment Analysis (ASA) in Dialectal Arabic (DA). Classical Arabic (CA), which is used in the Quran; Modern Standard Arabic (MSA), which is common in newspapers and education; and Dialectical Arabic (DA), which is used informally in chat rooms and social media platforms. Different Arab countries have various dialectical Arabic forms [22]. Thus, there are issues such as a shortage of Dialectical Arabic (DA) datasets and lexicons, stringent processes for gaining permission to reuse aggregated data, most current corpora not permitting free entry, and Dialectical Arabic analysis requiring a native speaker. As a result, the researchers want these gaps to be filled by developing a Saudi corpus and lexicon that can be used for data mining in the telecom industry. The researchers constructed a tweet generator that generates tweets automatically (the tweet includes a link to the questionnaire) using Python for all 20,000 individuals whose tweets we had previously compiled, but only 200 people participated. The tweet generator was built using Python code to generate tweets containing two elements: A link to the survey and mentions of participants' Twitter accounts [22]. A study focused on an experiment on two separate datasets extracted from Twitter, one with tweets in English only and the other with 23 different languages with 85% accuracy. This experiment examines how Twitter data shows society's thoughts and ideas on the study argued on social media. By adopting the following methods, multiple machine learning and natural language processing-based algorithms classify the most suitable model in how much it improves the performance and implement a Python library for sentiment analysis of an Altimetric dataset. Furthermore, they use machine learning algorithms with two existing baseline tools SentiStrength and Sentiment140, to carefully distinguish if the tweet includes positive or negative terms extracted from the tweeted papers. Finally, they achieved higher accuracy by comparing NLP-based methods to ML-based techniques [23].

Natalia Loukachevitch and Yuliya Rubtsova published a paper summarizing the outcomes of the reputation-oriented Twitter challenge done as part of the SentiRuEval evaluation of Russian sentiment-analysis systems. The examination includes tweets from two domains: telecom providers and banks. The goal was to determine whether a tweet's author had a positive or negative opinion toward a corporation referenced in the message. For example, tweets on a firm's reputation may express an author's viewpoint or positive or alarming facts about the company. This testing involved a total of nine teams, with most participants employing various machine-learning techniques. The primary purpose is to examine the existing situation and issues with the participants' methods. For example, the paper cited problems such as frequent misprints in tweets written on mobile phones, like "recreation abea" instead of "recreation area." In addition, there is a lot of slang, word contractions, and abbreviations in tweets [24]. In this paper, the authors present a brief description of the task, data, guidelines for the study, and results acquired by participants and approaches and analyze the problems with current systems. According to the participants' results, the authors' best-achieved performance in the reputation-oriented task for a particular domain relates to the difference between word probability distributions overtraining and test collections in this domain. Such disparities can occur in the reputation task because of recent spectacular events. Furthermore, at this time, integrating a general sentiment vocabulary and a general vocabulary of connotative words into machine-learning algorithms can significantly enhance results in the tweet reputation job. Also, they found that most participants could solve the available task of tweet classification; entity-oriented approaches did not yield better outcomes [24].

Twitter is considered as a valuable resource for sentiment analysis, as people use it to express their opinions of various topics. As Twitter is one of the most used social media in the Arab world, which can count as a platform that help with capturing the sentiments of the society especially on issues and social topics. This paper focus in unemployment in Saudi Arabia and the challenges the researcher may face when using sentiment analysis to capture the emotions and opinions in Arabic and offerings a method for organizing linguistic pre-processing to address the complication of handling the microblog format of tweets, as well as problems rendering non-standard Arabic dialects by the help of Naive Bayes (NB) and Support Vector Machine (SVM) algorithms. The Arabic language is known as a language that have a complex morphology, Roots of three, four, or five letters are used to create Arabic words. Prefixes, infixes, and suffixes are used to create words. Rooting and light stemming are two approaches for analyzing the morphology of Arabic words. Rooting is the process of removing prefixes, suffixes, and infixes from a word before converting it to its root form. On the other hand, light stemming just eliminates prefixes and suffixes [25].

Using social media gives people the chance to express their feeling and opinion about a specific situation. Social media applications work on real-time data. In the pandemic of Covid-19, people could share their opinions about the situation through these applications. This research will focus on hidden sentiments under people's reactions. Using Natural Language Processing and Machine Learning, we can analyze the data we collect from Twitter. Also, using Sentiment Analysis will help us to detect the text polarity and arrange it into three groups which are Positive, Negative, and Natural. This will help to get a general idea about people's feelings about the pandemic of Covid-19. Because human language is complex for computers to understand, Natural Language Processing is required. Sentiment analysis employs a variety of Natural Language Processing methodologies and algorithms [26].

### 3.3 Machine Learning and lexicon-based approaches

The paper proposed a 'HILATSA' system hybrid approach that combines machine learning and lexicon-based approaches. HILATSA contains three main parts. The first is an existing emotion lexicon, one built for the most popular phrases and one lexicon using several datasets to separate the words into positive, negative, and natural. The second is a trained and tested classifier. Finally, the semi-automatic learning part updates the system to recent language changes. For all the six used datasets, only the positive, negative, and neutral tweets are considered. They remove the mixed or the ones that cannot be identified. The suggested approach reached an accuracy of 73.67% and 83.73% [27]. Al-Twairesh et al. [28] applied the Sentiment Analysis on Saudi tweets written in Arabic. They classified the tweets into positive, negative, neutral, mixed, and intermediate. The intermediate category is for tweets with no clear results to be excluded.

In the pre-processing stage they removed tweets with media or URL, because most of them are spam. Also, they removed hashtag symbols, retweets, mentions, and tweets that include words from other languages. Furthermore, some Arabic letters normalized into one form. For example, the letter “ة” changed to “ه”.

### 3.4 Deep Learning and Bert based models

After releasing BERT, many research papers selected Bert for their Arabic sentiment analysis experiment, and the outcomes indicate that BERT has enormously enhanced the performance of models in sentiment analysis. Most studies are focused on binary classification, and there is some research on multiclass classification. In addition, there is research focused on sarcasm and sentiment detection in Arabic [29] they worked on seven BERT-Based models to determine the sentiment of a tweet or detect if a tweet is sarcasm. They achieved the best result using the MARBERT model. With the difficulty of detecting sarcasm, fine-tuning BERT and a Sentence- BERT are used to detect sarcasm [30]. Wadhawan proposed a Sarcasm and Sentiment Detection strategy by applying the released ArSarcasm-v2 dataset and processing it into a better format for the models. Then train and test various Bert models with the resulting data to determine which model achieves the most acceptable result for the ArSarcasm-v2 dataset [31]. Arabic monolingual BERT models performed better than BERT multilingual models. Husain and Uzuner Applied several Arabic offensive datasets in the AraBERT model. They found that combining data from multiple datasets lowers the model's performance compared to testing it on individual datasets [32].

Table 2 summarizes the reviewed literature in terms of language, dataset size and classes, approach used and the accuracy achieved by each technique. It is apparent that there is room for improvement especially in Arabic language case.

**Table 2.** Summary of literature review

| Ref. | Dataset | | | Approach | Result (Acc.) |
|---|---|---|---|---|---|
| | Lang. | No. of tweets | Classes/ Imbalance? | | |
| [18] | Arabic | - | Multi/ Yes | Ensemble model LSTM and CNN | 64.46% |
| [19] | English | 170,000 | Multi/ Yes | LSTM and CNN | 86.07% |
| [20] | English and Malay | 83,647 | Binary/ Yes | LSTM and CNN | 93.78%. |
| [21] | English | <10,000 | Multi/ Yes | LSTM and BiRNN | - |
| [22] | Arabic | 20,000 | Binary/ Yes | ML + NLP | 91% |
| [23] | English and 23 different languages | 6388 | Multi/ Yes | ML + NLP | 85% |
| [24] | Russian | - | Multi/ Yes | ML + NLP | - |
| [25] | Arabic | - | Multi/ Yes | ML + NLP | - |
| [26] | English | - | Multi/ Yes | ML + NLP | - |
| [27] | Arabic | 41,041 | Multi/ both types | ML and lexicon-based approaches | 73.67% and 83.73% |
| [28] | Arabic | 17,573 | Multi/ Yes | ML and lexicon-based approaches | - |

## 4. METHODOLOGY

In this section, we first briefly describe the different data preprocessing steps and background about the model we used in the experiments.

### 4.1 Preprocessing

To clean up a tweet, the Tweets Cleaning Function is used. This is mainly to remove all hash-tags, usernames, and URLs. As these are the extra tokens that do not incorporate in the proposed model as such. Also, it should remove extra spaces, extra symbols, and multiple characters treated as noise characters.

Moreover, another function was used to eliminate all duplicate data. Such as while downloading the file, it happens same tweets are repeated etc. Finally, there was a process of transforming the letters that appeared in different forms into a single form. Since in the Arabic language, such letters may take several forms. Hence, for sake of uniformity and simplicity in the Arabic tweet text, this process was performed. For example, {أآإ} was replaced with {ا}, {ئ} with {ي}, and {ؤ} with {و}.

### 4.2 Background

Recent years have seen a massive explosion in the development of bidirectional transformer-based models, particularly for Arabic. They serve as powerful tools for transfer learning that help enhance the performance of natural language processing (NLP). MARBERT will be tested in our experiment as BERT-based models that help achieve state-of-the-art results on various downstream NLP tasks. Abdul-Mageed at el. [33] developed MARBERT, which is based on the BERT-based model, except that it does not use the next sentence prediction (NSP) objective because it was trained on short tweets.

Also, MARBERT trained on both Dialectal (DA) and Modern Standard Arabic (MSA) tweets and has 15.6B tokens. The prediction is based on the input tokens fed to the encoders by considering the output label. We will train the MARBERT model in both datasets. Figure 1 shows the general structure of BERT based model while mathematical foundation can be found in reference [34].

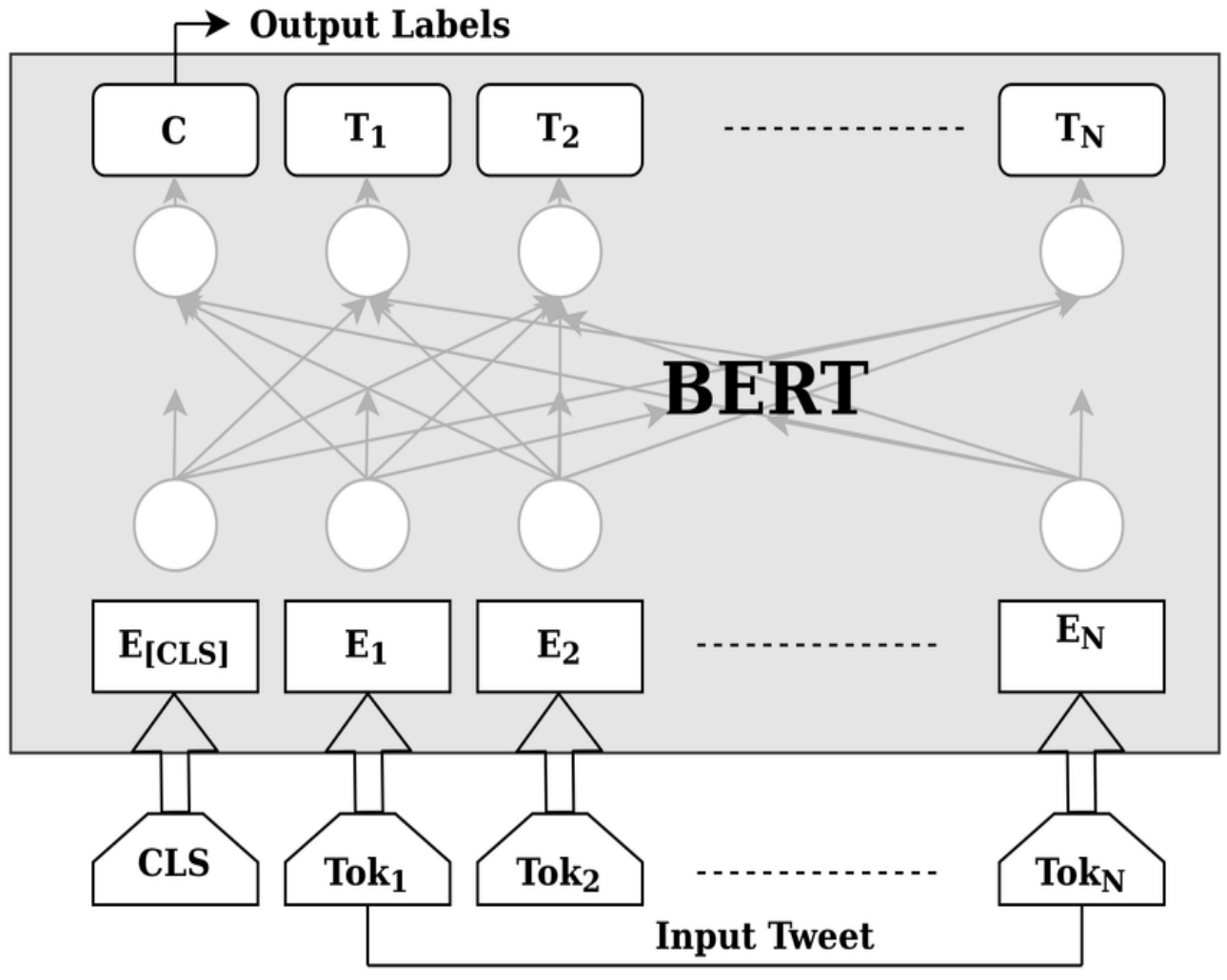

**Figure 1.** General structure of BERT based model

## 5. EXPERIMENT AND ANALYSIS

### 5.1 Dataset

The client privately provides the two datasets. The first one contains STC customer feedback collected from Twitter only and more than twenty-two thousand (22K) tweets labeled manually (Negative, Positive, Neutral, Sarcasm, and Indeterminate).

The second one is the spam tweets dataset containing more than twenty-four (24K) tweets with some unwanted data labeled as (Yes, NO) to indicate whether it is spam or not. The labeling in pre-processed dataset was made on behalf of a carved word-net containing potentially the unwanted keywords and phrases.

### 5.2 Experiment

As shown in Figure 2, the proposed architecture is mainly based on the MARBERT model. Our experiment will focus on spam and sentiment detection in Arabic, we first trained the model with the spam tweets dataset, and. Then we trained the model with sentiment tweets dataset. Many earlier studies have concentrated on model performance on a balanced dataset.

In practice, however, the distribution of samples from various classes is often unbalanced. For example, Twitter users may prefer to share negative emotions in sarcasm rather than in a normal way. As a result, some emotions arise more frequently than others, resulting in poor classification ability in terms of classifier's tendency to result in the dominant class. Accordingly, in our experiment, we noticed that the model was not learning the (indeterminate and sarcasm) tweets adequately, so we decided to collect more and diverse type of data from Twitter.

For example, the sarcasm was initially 200, and eventually it increased to 1,400. As a result, the performance improved the model training scores increased. Another way to get rid of the same issue is the use of some balancing technique like Balanced Bagging etc. Consequently, dataset becomes balanced and provides better and fair results. As demonstrated in the subsequent sections.

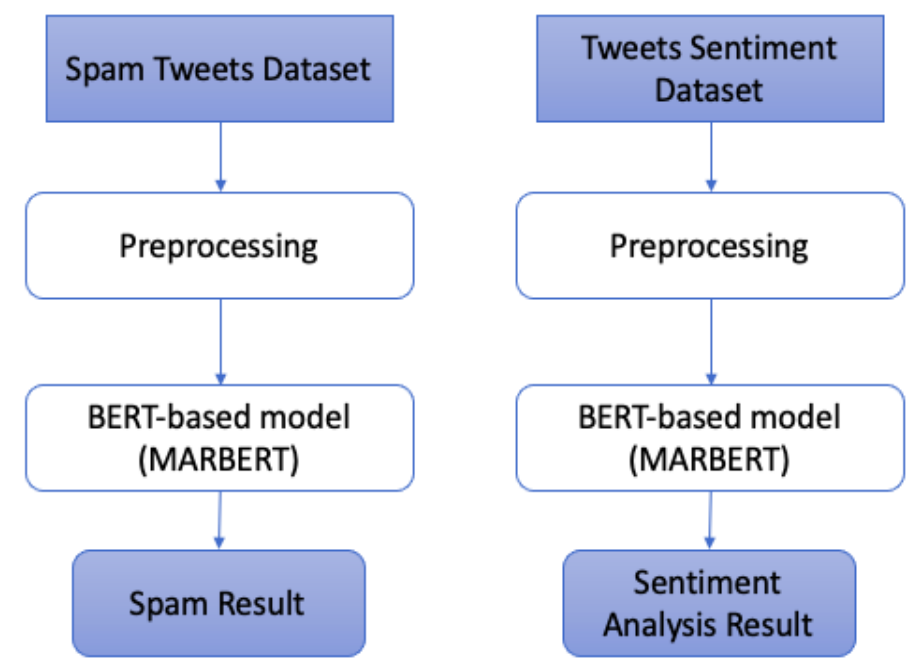

**Figure 2.** Architecture based on MARBERT

### 5.3 Oversampling and under-sampling

Oversampling and under-sampling are standard strategies for dealing with unbalanced datasets. Although they are not the best solutions to balance the data, an oversampling method merely samples a part of it repeatedly. Even though the data appears to be balanced, no new data is introduced, which increases the impact of noise on the model. Furthermore, because under-sampling might result in considerable data loss, it is not a possible option for a small dataset. So, we decide not to use this solution.

### 5.4 Spam detection

When we first trained the model with the spam tweets dataset, we could successfully obtain good results in Table 3. In addition, the table presents a review of recent works on spam detection models and proposed technique outperforms.

**Table 3.** F1-score of spam tweets with review of recent studies

| Author | Classifier | F1-score | Dataset |
|---|---|---|---|
| Saab et al. [15] | SVM, NB, LMSVM, Decision tree, ANN | 0.93 achieved by SVM | Spam base contains 4597 rows |
| Yaseen [16] | Bert Based | 0.96 | Spam base contains 11,297 rows |
| MARBERT | Bert Based | 0.98 | spam tweets dataset containing more than 24,000 rows |

### 5.5 Evaluation

Precision, recall, F1-score, and support are commonly used as evaluation indices for sentiment analysis tasks, so we employed these metrics to assess our algorithms' performance in our experiment. The precision, recall, and F1-score of each emotion are calculated as shown in the following formulas, respectively [35-40].

$$\begin{aligned} Precision &= \frac{TP}{TP+FP} \\ Recall &= \frac{TP}{TP+FN} \\ F1-score &= \frac{2\times \text{ Precision } \times \text{ Recall}}{\text{Precision } + \text{ Recall}} \end{aligned} \tag{1}$$

### 5.6 Fine tuning

We consider different hyper-parameter tuning to achieve the most suitable performance [41-50]. Experiments with various batch sizes, learning rates, and epoch counts have been performed. We looked for the most useful fine-tuned settings for unbalanced and small datasets and the fine-tuned settings that gave the previous researchers the highest scores. The result in Table 4 and Table 5, respectively, shows that using a batch size of 96 produces better results than a batch size of 16 or 14. In terms of the learning rate, a lower learning rate will cause training to converge gradually, but it will increase the model's performance. The 2e-5 learning rate produces more significant results than the other learning rates. The 265 Maximum Sequence Length yields more significant results than the others. Furthermore, we chose 4 and 5 as epoch parameters. According to the experiment, the best parameters of MARBERT for our work are a train batch size of 96, an evaluation batch size of 32, a learning rate of 2e-5 (difference between predicted and actual value), and 5 epochs, respectively.

**Table 4.** Fine-Tuning and F1-score

| Epsilon (Adam optimizer) | Learning rate | Max Seq. Len. | Epoch | Batch sizes | F1-score |
|---|---|---|---|---|---|
| 1e-8 | 5e-5 | 125 | 2 | 16-8 | 0.62 |
| 1e-8 | 1e-5 | 125 | 4 | 40-14 | 0.65 |
| 2e-5 | 1.215e-05 | 125 | 4 | 32-16 | 0.69 |
| 1e-8 | 2e-5 | 125 | 4 | 64-32 | 0.70 |
| 1e-8 | 2e-5 | 128 | 4 | 64-32 | 0.73 |
| 1e-8 | 2e-5 | 256 | 5 | 96-32 | 0.75 |

**Table 5.** Detailed F1-score of each class

| Batch sizes | -ive | +ive | Neutral | Sarcasm | Indeterminate |
|---|---|---|---|---|---|
| 16 | 0.80 | 0.85 | 0.79 | 0.33 | 0.50 |
| 96 | 0.85 | 0.92 | 0.89 | 0.54 | 0.56 |

## 6. CONCLUSIONS

The study presents a BERT-based model for multilabel sentiment analysis in unbalanced Arabic data from Twitter. We used MARBERT model and experimented with different parameters. Considering the unbalanced dataset's characteristics, we collected more data from Twitter and evaluated various hyperparameter tuning options. According to the experiment, we found that the best parameters of MARBERT for our work are a train batch size of 96; an evaluation batch size of 32, a learning rate of 2e-5, and 5 epochs is enough for our data. The parameters were obtained after several trials. Even though we use a public BERT pre-trained model, the result is still considerable for unbalanced data. The experiment shows that MARBERT performs well in a multilabel Arabic sentiment analysis problem. In future, deep learning, fusion based, transfer learning and federated learning-based approaches can be used in contrast to the cloud computing paradigms [51-76].

## ACKNOWLEDGMENT

The authors would like to acknowledge SAUDI ARAMCO Cybersecurity Chair at College of Computer Science and Information Technology (CCSIT), Imam Abdulrahman Bin Faisal University (IAU), Dammam, Kingdom of Saudi Arabia for supporting and funding this research.

## REFERENCES

[1] Horev, R. (2018). BERT Explained: State of the art language model for NLP. Towards Data Science, 10.

[2] Alqarni, A., Rahman, A. (2023) Arabic Tweets-based Sentiment Analysis to investigate the impact of COVID-19 in KSA: A deep learning approach. Big Data and Cognitive Computing, Preprint.

[3] Abuzayed, A., Al-Khalifa, H. (2021). Sarcasm and sentiment detection in Arabic tweets using BERT-based models and data augmentation. In Proceedings of the

Sixth Arabic Natural Language Processing Workshop, 312-317.
[4] Bashmal, L., AlZeer, D. (2021). ArSarcasm shared task: An ensemble BERT model for Sarcasm Detection in Arabic tweets. In Proceedings of the Sixth Arabic Natural Language Processing Workshop, 323-328.
[5] Wadhawan, A. (2021). Arabert and farasa segmentation-based approach for sarcasm and sentiment detection in arabic tweets. arXiv preprint arXiv:2103.01679. https://doi.org/10.48550/arXiv.2103.01679
[6] Husain, F., Uzuner, O. (2021). Transfer learning approach for Arabic offensive language detection System--BERT-Based model. arXiv preprint arXiv:2102.05708. https://doi.org/10.48550/arXiv.2102.05708
[7] Al-Twairesh, N. (2021). The evolution of language models applied to emotion analysis of Arabic tweets. Information, 12(2): 84. https://doi.org/10.3390/info12020084
[8] Talafha, B., Ali, M., Za'ter, M.E., Seelawi, H., Tuffaha, I., Samir, M., Al-Natsheh, H.T. (2020). Multi-dialect Arabic Bert for country-level dialect identification. arXiv preprint arXiv:2007.05612. https://doi.org/10.48550/arXiv.2007.05612
[9] Abdelali, A., Hassan, S., Mubarak, H., Darwish, K., Samih, Y. (2021). Pre-training Bert on Arabic tweets: Practical considerations. arXiv preprint arXiv:2102.10684. https://doi.org/10.48550/arXiv.2102.10684
[10] Alturayeif, N., Luqman, H. (2021). Fine-grained sentiment analysis of Arabic COVID-19 tweets using BERT-based transformers and dynamically weighted loss function. Applied Sciences, 11(22): 10694. https://doi.org/10.3390/app112210694
[11] Habbat, N., Anoun, H., Hassouni, L. (2021). A novel hybrid network for arabic sentiment Analysis using fine-tuned AraBERT model. International Journal on Electrical Engineering and Informatics, 13(4): 801-812. https://doi.org/10.15676/ijeei.2021.13.4.3
[12] Antoun, W., Baly, F., Hajj, H. (2020). Arabert: Transformer-based model for Arabic language understanding. arXiv preprint arXiv:2003.00104. https://doi.org/10.48550/arXiv.2003.00104
[13] Alnajrani, B., Alghamdi, A., Alotaibi, M., Aldawod, S., Rahman A., Nabil, M. (2022). A novel approach to Wikipedia references classification. ICIC Express Letters, 13(12): 1-9. http://dx.doi.org/10.24507/icicelb.13.12.1321
[14] Safaya, A., Abdullatif, M., Yuret, D. (2020). Kuisail at semeval-2020 task 12: Bert-CNN for offensive speech identification in social media. In Proceedings of the Fourteenth Workshop on Semantic Evaluation, 2054-2059.
[15] Saab, S.A., Mitri, N., Awad, M. (2014). Ham or spam? A comparative study for some content-based classification algorithms for email filtering. In MELECON 2014-2014 17th IEEE Mediterranean Electrotechnical Conference, 339-343. https://doi.org/10.1109/MELCON.2014.6820574
[16] Yaseen, Q. (2021). Spam email detection using deep learning techniques. Procedia Computer Science, 184: 853-858. https://doi.org/10.1016/j.procs.2021.03.107
[17] Musleh, D.A., Alkhales, T.A., Almakki, R.A., Alnajim, S.E., Almarshad, S.K., Alhasaniah, R.S., Almuqhim, A.A. (2022). Twitter Arabic sentiment analysis to detect depression using machine learning. CMC-Computers Materials & Continua, 71(2): 3463-3477. https://doi.org/10.32604/cmc.2022.022508
[18] Heikal, M., Torki, M., El-Makky, N. (2018). Sentiment analysis of Arabic tweets using deep learning. Procedia Computer Science, 142: 114-122. https://doi.org/10.1016/j.procs.2018.10.466
[19] Hanafy, M., Khalil, M.I., Abbas, H.M. (2018). Combining classical and deep learning methods for Twitter sentiment analysis. In IAPR Workshop on Artificial Neural Networks in Pattern Recognition, 281-292. https://doi.org/10.1007/978-3-319-99978-4_22
[20] Feizollah, A., Ainin, S., Anuar, N.B., Abdullah, N.A.B., Hazim, M. (2019). Halal products on Twitter: Data extraction and sentiment analysis using stack of deep learning algorithms. IEEE Access, 7: 83354-83362. https://doi.org/10.1109/ACCESS.2019.2923275
[21] Jiang, L., Suzuki, Y. (2019). Detecting hate speech from tweets for sentiment analysis. In 2019 6th International Conference on Systems and Informatics (ICSAI), 671-676. https://doi.org/10.1109/ICSAI48974.2019.9010578
[22] Almuqren, L., Cristea, A. (2021). AraCust: A Saudi telecom tweets corpus for sentiment analysis. PeerJ Computer Science, 7: e510. http://dx.doi.org/10.7717/peerj-cs.510
[23] Hassan, S.U., Saleem, A., Soroya, S.H., Safder, I., Iqbal, S., Jamil, S., Nawaz, R. (2021). Sentiment analysis of tweets through Altmetrics: A machine learning approach. Journal of Information Science, 47(6): 712-726. https://doi.org/10.1177/0165551520930917
[24] Král, P., Matoušek, V. (2015). Text, speech, and dialogue: 18th International Conference, TSD 2015 Pilsen, Czech Republic, 2015: 551-559.
[25] Alwakid, G., Osman, T., Hughes-Roberts, T. (2017). Challenges in sentiment analysis for Arabic social networks. Procedia Computer Science, 117: 89-100. https://doi.org/10.1016/j.procs.2017.10.097
[26] Khan, R., Shrivastava, P., Kapoor, A., Tiwari, A., Mittal, A. (2020). Social media analysis with AI: sentiment analysis techniques for the analysis of Twitter covid-19 data. Critical Rev, 7(9): 2761-2774.
[27] Elshakankery, K., Ahmed, M.F. (2019). HILATSA: A hybrid Incremental learning approach for Arabic tweets sentiment analysis. Egyptian Informatics Journal, 20(3): 163-171. https://doi.org/10.1016/j.eij.2019.03.002
[28] Al-Twairesh, N., Al-Khalifa, H., Al-Salman, A., Al-Ohali, Y. (2017). Arasenti-tweet: A corpus for Arabic sentiment analysis of Saudi tweets. Procedia Computer Science, 117: 63-72. https://doi.org/10.1016/j.procs.2017.10.094.
[29] Warda, L., Kaladzavi, G., Samdalle, A., Kolyang. (2022). Integration of ontology transformation into hidden Markov model. Information Dynamics and Applications, 1(1): 2-13. https://doi.org/10.56578/ida010102
[30] Musleh, D., Halawani, K., Mahmoud, S. (2015) Fuzzy modeling for handwritten Arabic numeral recognition. International Arab Journal of Information Technology, 14(4): 1-10.
[31] Farooq, M.S., Abbas, S., Rahman A., Sultan, K., Khan, M.A., Mosavi, Amir. (2023). A fused machine learning approach for intrusion detection system. Computers, Materials & Continua, 74(2): 2607-2623.
[32] Qureshi M.A., Asif, M., Anwar, S., Shaukat, U., Rahman,

A., Khan, M.A., Mosavi, A. (2023). Aspect level songs rating based upon reviews in English. Computers, Materials & Continua, 74(2): 2589-2605.
[33] Abdul-Mageed, M., Elmadany, A., Nagoudi, E.M.B. (2020). ARBERT & MARBERT: Deep bidirectional transformers for Arabic. arXiv [cs.CL].
[34] Vaswani, A., Shazeer, N., Parmar, N., Uszkoreit, J., Jones, L., Gomez, A.N., Polosukhin, I. (2017). Attention is all you need. Advances in neural information processing systems, Part of Advances in Neural Information Processing Systems 30 (NIPS 2017), 30: 6000-6010.
[35] Ur, A., Rahman, S., Naseer, I., Majeed, R., Musleh, D., Gollapalli, M.A.S., Khan, M.A. (2021). Supervised machine learning-based prediction of covid-19. Computers, Materials and Continua, 69(1): 21-34. https://doi.org/10.32604/cmc.2021.013453
[36] Alotaibi, S.M., Basheer, M.I., Khan, M.A. (2021). Ensemble machine learning based identification of pediatric epilepsy. Computers, Materials & Continua, 68(1): 149-165. https://doi.org/10.32604/cmc.2021.015976
[37] Zaman, G., Mahdin, H., Hussain, K., Abawajy, J., Mostafa, S.A. (2021). An ontological framework for information extraction from diverse scientific sources. IEEE Access, 9: 42111-42124. https://doi.org/10.1109/ACCESS.2021.3063181
[38] Dash, S., Luhach, A.K., Chilamkurti, N., Baek, S., Nam, Y. (2019). A Neuro-fuzzy approach for user behaviour classification and prediction. Journal of Cloud Computing, 8(1): 1-15. https://doi.org/10.1186/s13677-019-0144-9
[39] Rahman, A. (2019). Memetic computing based numerical solution to Troesch problem. Journal of Intelligent & Fuzzy Systems, 37(1): 1545-1554. https://doi.org/10.3233/JIFS-18579
[40] Rahman, A. (2019). Optimum information embedding in digital watermarking. Journal of Intelligent & Fuzzy Systems, 37(1): 553-564. https://doi.org/10.3233/JIFS-162405
[41] Musleh, D., Ahmed, R., Alhaidari, F. (2019). A novel approach to Arabic keyphrase extraction. ICIC Express Letters. Part B, Applications: An International Journal of Research and Surveys, 10(10): 875-884. https://doi.org/10.24507/icicelb.10.10.875
[42] Ahmad, M., Farooq, U., Atta-Ur-Rahman, Alqatari, A., Dash, S., Luhach, A.K. (2019). Investigating TYPE constraint for frequent pattern mining. Journal of Discrete Mathematical Sciences and Cryptography, 22(4): 605-626. https://doi.org/10.1080/09720529.2019.1637158
[43] Rahman, A. (2020). GRBF-NN based ambient aware real time adaptive communication in DVB-S2. Journal of Ambient Intelligence and Humanized Computing, 1-11. https://doi.org/10.1007/s12652-020-02174-w
[44] Alhaidari, F., Rahman, A., Zagrouba, R. (2020). Cloud of things: architecture, applications and challenges. Journal of Ambient Intelligence and Humanized Computing, 1-19. https://doi.org/10.1007/s12652-020-02448-3.
[45] Rahman, A.U., Dash, S., Luhach, A.K. (2021). Dynamic MODCOD and power allocation in DVB-S2: A hybrid intelligent approach. Telecommunication Systems, 76(1): 49-61. https://doi.org/10.1007/s11235-020-00700-x
[46] Dash, S., Ahmad, M., Iqbal, T. (2021). Mobile cloud computing: A green perspective. In Intelligent Systems, 185: 523-533. https://doi.org/10.1007/978-981-33-6081-5_46
[47] Alhaidari, F., Almotiri, S.H., Al Ghamdi, M.A., Khan, M.A., Rehman, A., Abbas, S., Khan, K.M. (2021). Intelligent software-defined network for cognitive routing optimization using deep extreme learning machine approach. Computers, Materials & Continua, 67(1): 1269-1285. https://doi.org/10.32604/cmc.2021.013303
[48] Rahman, A.U., Alqahtani, A., Aldhafferi, N., Nasir, M.U., Khan, M.F., Khan, M.A., Mosavi, A. (2022). Histopathologic oral cancer prediction using oral squamous cell carcinoma biopsy empowered with transfer learning. Sensors, 22(10): 3833. https://doi.org/10.3390/s22103833
[49] Rahman, A.U., Abbas, S., Gollapalli, M., Ahmed, R., Aftab, S., Ahmad, M., Mosavi, A. (2022). Rainfall prediction system using machine learning fusion for smart cities. Sensors, 22(9): 3504. https://doi.org/10.3390/s22093504
[50] Ghazal, T.M., Al Hamadi, H., Umar Nasir, M., Gollapalli, M., Zubair, M., Adnan Khan, M., Yeob Yeun, C. (2022). Supervised machine learning empowered multifactorial genetic inheritance disorder prediction. Computational Intelligence and Neuroscience, 2022: Article ID: 1051388. https://doi.org/10.1155/2022/1051388
[51] Alhaidari, F., Shaib, N.A., Alsafi, M., Alharbi, H., Alawami, M., Aljindan, R., Zagrouba, R. (2022). ZeVigilante: Detecting zero-day malware using machine learning and sandboxing analysis techniques. Computational Intelligence and Neuroscience, 2022: Article ID: 1615528. https://doi.org/10.1155/2022/1615528
[52] Ibrahim, N.M., Gabr, D.G.I., Rahman, A.U., Dash, S., Nayyar, A. (2022). A deep learning approach to intelligent fruit identification and family classification. Multimedia Tools and Applications, 81: 27783-27798. https://doi.org/10.1007/s11042-022-12942-9
[53] Gollapalli, M., Musleh, D., Ibrahim, N., Khan, M.A., Abbas, S., Atta, A., Omer, A. (2022). A neuro-fuzzy approach to road traffic congestion prediction. Computers, Materials and Continua, 72(3): 295-310. https://doi.org/10.32604/cmc.2022.027925
[54] Dilawari, A., Khan, M.U.G., Al-Otaibi, Y.D., Rehman, Z.U., Rahman, A.U., Nam, Y. (2021). Natural language description of videos for smart surveillance. Applied Sciences, 11(9): 3730. https://doi.org/10.3390/app11093730
[55] Zagrouba, R., Khan, M.A., Atta Ur, R., Muhammad Aamer, S., Muhammad Faheem, M., Rehman, A., Muhammad Farhan, K. (2021). Modelling and simulation of COVID-19 outbreak prediction using supervised machine learning. Computers, Materials, & Continua, 66(3): 2397-2407.
[56] Khan, M.A., Abbas, S., Atta, A., Ditta, A., Alquhayz, H., Khan, M.F., Naqvi, R.A. (2020). Intelligent cloud-based heart disease prediction system empowered with supervised machine learning. Computers, Materials & Continua, 65(1): 139-151. https://doi.org/10.32604/cmc.2020.011416
[57] Naseem, M.T., Qureshi, I.M., Muzaffar, M.Z. (2020). Robust and fragile watermarking for medical images

using redundant residue number system and chaos. Neural Network World, 30(3): 177-192. https://doi.org/10.14311/nnw.2020.30.013

[58] Nasir, M.U., Zubair, M., Ghazal, T.M., Khan, M.F., Ahmad, M., Rahman, A.U., Mansoor, W. (2022). Kidney cancer prediction empowered with blockchain security using transfer learning. Sensors, 22(19): 7483. https://doi.org/10.3390/s22197483

[59] AlKhulaifi, D., AlQahtani, M., AlSadeq, Z., Rahman, A., Musleh, D. (2022). An overview of self-adaptive differential evolution algorithms with mutation strategy. Mathematical Modelling of Engineering Problems, 9(4): 1017-1024. https://doi.org/10.18280/mmep.090419

[60] Rahman, A., Musleh, D., Nabil, M., Alubaidan, H., Gollapalli, M., Krishnasamy, G., Almoqbil, D., Khan, M.A.A., Farooqui, M., Ahmed, M.I.B., Ahmed, M.S., Mahmud, M. (2022). Assessment of information extraction techniques, models and systems. Mathematical Modelling of Engineering Problems, 9(3): 683-696. https://doi.org/10.18280/mmep.090315

[61] ur Rahman, A. (2022). Geo-Spatial disease clustering for public health decision making. Informatica, 46(6): 3827. https://doi.org/10.31449/inf.v46i6.3827

[62] Khan, M.A.A., AlAyat, M., AlGhamdi, J., AlOtaibi, S. M., AlZahrani, M., AlQahtani, M., Jan, F. (2023). WeScribe: An intelligent meeting transcriber and analyzer application. In Proceedings of Third International Conference on Computing, Communications, and Cyber-Security, 755-766. https://doi.org/10.1007/978-981-19-1142-2_59

[63] Rahman, A.U., Asif, R.N., Sultan, K., Alsaif, S.A., Abbas, S., Khan, M.A., Mosavi, A. (2022). ECG classification for detecting ECG arrhythmia empowered with deep learning approaches. Computational intelligence and neuroscience, 2022: Article ID 6852845. https://doi.org/10.1155/2022/6852845

[64] Rahman, A.U., Nasir, M.U., Gollapalli, M., Alsaif, S.A., Almadhor, A.S., Mehmood, S., Mosavi, A. (2022). IoMT-based mitochondrial and multifactorial genetic inheritance disorder prediction using machine learning. Computational Intelligence and Neuroscience, 2022: Article ID 2650742. https://doi.org/10.1155/2022/2650742

[65] Nasir, M.U., Khan, S., Mehmood, S., Khan, M.A., Rahman, A.U., Hwang, S.O. (2022). IoMT-Based osteosarcoma cancer detection in histopathology images using transfer learning empowered with blockchain, fog computing, and edge computing. Sensors, 22(14): 5444. https://doi.org/10.3390/s22145444.

[66] Abbas, S., Raza, S.A., Khan, M.A., Rahman, A., Sultan, K., Mosavi, A. (2023). Automated file labeling for heterogeneous files organization using machine learning. Computers. Materials & Continua 74(2): 3263-3278.

[67] Mahmud, M., Rahman, A., Lee, M., Choi, J.Y. (2020). Evolutionary-based image encryption using RNA codons truth table. Optics & Laser Technology, 121: 105818.

[68] Gollapalli, M. (2022). Ensemble Machine Learning Model to Predict the Waterborne Syndrome. Algorithms, 15(3): 93.

[69] Rahman, A., Mahmud, M., Iqbal, T., Saraireh, L., Kholidy, H., Gollapalli, M., Musleh, D., Alhaidari, F., Almoqbil, D., Ahmed, M.I.B. (2022). Network anomaly detection in 5G networks. Mathematical Modelling of Engineering Problems, 9(2): 397-404. https://doi.org/10.18280/mmep.090213

[70] Ahmad, M., Qadir, M.A., Rahman, A. et al. (2020). Enhanced query processing over semantic cache for cloud based relational databases. J Ambient Intell Human Comput. https://doi.org/10.1007/s12652-020-01943-x

[71] Rahman, A., Qureshi, I.M., Malik, A.N., Naseem, M.T. (2016). Dynamic resource allocation in OFDM systems using DE and FRBS. Journal of Intelligent and Fuzzy Systems, 26(4): 2035-2046.

[72] Rahman, A., Alhaidari, F. (2018). Querying RDF Data. Journal of Theoretical and Applied Information Technology, 26(22): 7599-7614.

[73] Sajid, N.A., Ahmad, M., Afzal, M.T., Rahman, A. (2021). Exploiting papers' reference's section for multi-label computer science research papers' classification. Journal of Information & Knowledge Management 20(2): 1-21.

[74] Sajid, N.A., Ahmad, M., Rahman, A., Zaman, G., Ahmed, M.S. (2023). A novel metadata based multi-label document classification technique. Computer Systems Science and Engineering, 46(1): 1-16.

[75] Rahman, A., Ibrahim, N.M., Musleh, D., Khan, M.A.A., Chabani, S., Dash, S. (2022). Cloud-based smart grids: Opportunities and challenges. In Proceedings Biologically Inspired Techniques in Many Criteria Decision Making, 1-13, Smart Innovation, Systems and Technologies, vol. 271, Springer, Singapore. https://doi.org/10.1007/978-981-16-8739-6_1

[76] Badi, I., Bouraima, M.B., Jibril, M.L. (2022). Risk assessment in construction projects using the grey theory. Journal of Engineering Management and Systems Engineering, 1(2): 58-66. https://doi.org/10.56578/jemse010203